\documentclass[11pt,a4paper]{llncs}
\usepackage{amsmath}
\usepackage{amssymb}
\usepackage{graphicx}
\usepackage{marvosym}
\usepackage{url}
\usepackage{fancyhdr}

\usepackage{geometry}
\newcommand{\keywords}[1]{\par\addvspace\baselineskip
\noindent\keywordname\enspace\ignorespaces#1}

\fancypagestyle{firstpage}{\fancyhf{}
}

\usepackage{tcolorbox}

\begin{document}


\title{\LARGE{Beyond Specialization: Assessing the Capabilities of MLLMs in Age and Gender Estimation}}


%
%
\author{\large{Maksim Kuprashevich \and
Grigorii Alekseenko \and
Irina Tolstykh}}

\institute{Layer Team, R\&D Department, SaluteDev \\
\email{\{mvkuprashevich,grigoriyalexeenko,irinakr4snova\}@gmail.com}}

%


%
%


\maketitle

\thispagestyle{firstpage}

\begin{abstract}
Multimodal Large Language Models (MLLMs) have recently gained immense popularity. Powerful commercial models like ChatGPT and Gemini, as well as open-source ones such as LLaVA, are essentially general-purpose models and are applied to solve a wide variety of tasks, including those in computer vision. These neural networks possess such strong general knowledge and reasoning abilities that they have proven capable of working even on tasks for which they were not specifically trained. We compared the capabilities of the most powerful MLLMs to date including ShareGPT4V, ChatGPT 4V/4O, and LLaVA Next in the specialized task of age and gender estimation, with the state-of-the-art specialized model MiVOLO384. In our study, we discovered that the fine-tuned open-source ShareGPT4V model is capable of outperforming the specialized model in age and gender estimation tasks. At the same time, the proprietary ChatGPT-4O beats both in the age estimation task but does not perform as confidently in gender recognition. This gives interesting insights about the strengths and weaknesses of the participating models and suggests that with a few tweaks, general-purpose MLLM models can match or even surpass specialized ones in certain fields. Even though these fine-tuned models might require more computing power, they offer big benefits for tasks where computing power is not a limiting factor and where the best accuracy is key, such as data annotation.

\keywords{MLLM, VLM, Human Attribute Recognition, Age estimation, Gender estimation, Large Models Generalization}
\end{abstract}

\section{Introduction}

\begin{figure}[t]
\centering
\includegraphics[width=10cm]{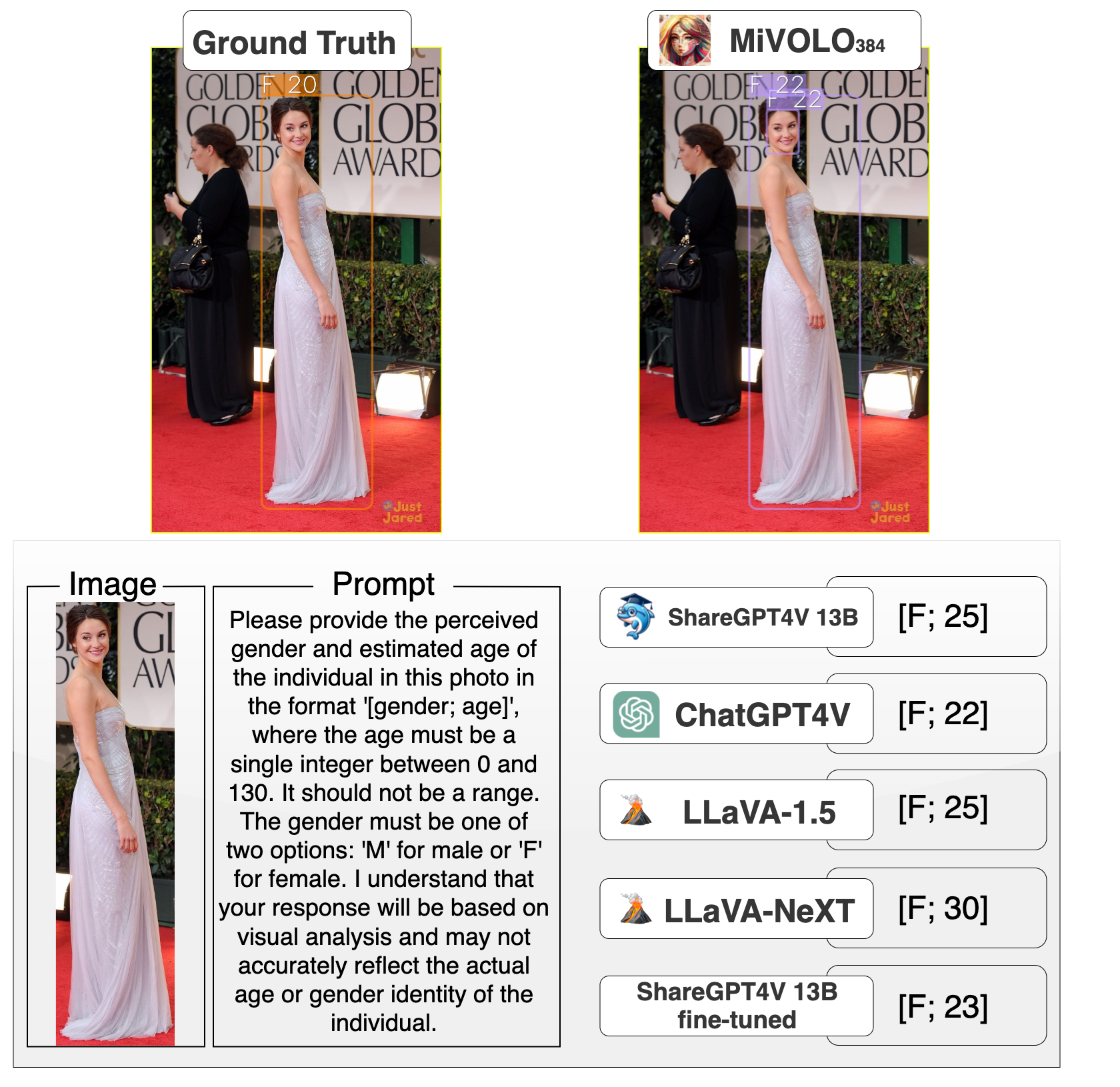}
\caption{An example of evaluated models predictions. The image illustrates output of specialized $MiVOLO_{384}$ model and different MLLMs. $MiVOLO_{384}$ makes predictions based on the face and body crops. Other models make predictions based on prompt and image of body crop.}\label{fig:example}
\end{figure}

The rapid development of multimodal large language models (MLLMs or LMMs) has been noteworthy, particularly those integrating language and vision modalities (LVMs). Their advancement is attributed to their high accuracy, generalization capability, reasoning skills, and robust performance with out-of-distribution data. These versatile models excel not only as AI assistants but also in handling unforeseen tasks beyond their initial training scope. The impact of MLLMs is profound, evolving so swiftly that it raises questions about the relevance of specialized models in certain areas. Moreover, there is an increasing interest in using MLLMs for specific computer vision tasks, such as object segmentation, and incorporating them into complex pipelines, such as instruction-based image editing.

We explored the competitiveness of MLLMs in the specific domain of age and gender estimation. Initially, we conducted preliminary tests with ChatGPT-4V \cite{chatgpt}. The results were highly encouraging, prompting a comprehensive evaluation of these neural networks' potential, including leading open-source solutions such as LLaVA \cite{llava,llava_improved} and ShareGPT-4V \cite{sharegpt}, which is also based on LLaVA. Later, we updated this work with results for ChatGPT-4O, the newest and most powerful OpenAI model available at the time.

We pursued several goals in these experiments:
\begin{itemize}
\item We aimed to compare the best general-purpose MLLMs with specialized models and understand their capability to replace them. Despite the huge difference in computational costs and speed, for some tasks, this is not crucial. This includes tasks such as labeling new data or filtering old datasets.
\item We were interested in what results could be achieved if an MLLM was fine-tuned for a specific task on a large target dataset. With the same motivations, many tasks require maximum accuracy and do not require fast inference.
\item Since the nature of specialized models and large general-purpose models is fundamentally different, it was reasonable to expect that such experiments could shed more light on the strengths and weaknesses of both approaches.
\end{itemize}
For the experiments, we tried to measure SOTA MLLM models: LLaVA 1.5 and LLaVA-NeXT, ShareGPT4V, and ChatGPT4. We were unable to measure the newly released Gemini Ultra, as it outright refused to work with images of people.

We've also made improvements to the state-of-the-art specialized model MiVOLO \cite{mivolo2023} to ensure fair competition among cutting-edge models.

\begin{figure}[t]
\centering
\includegraphics[height=5cm]{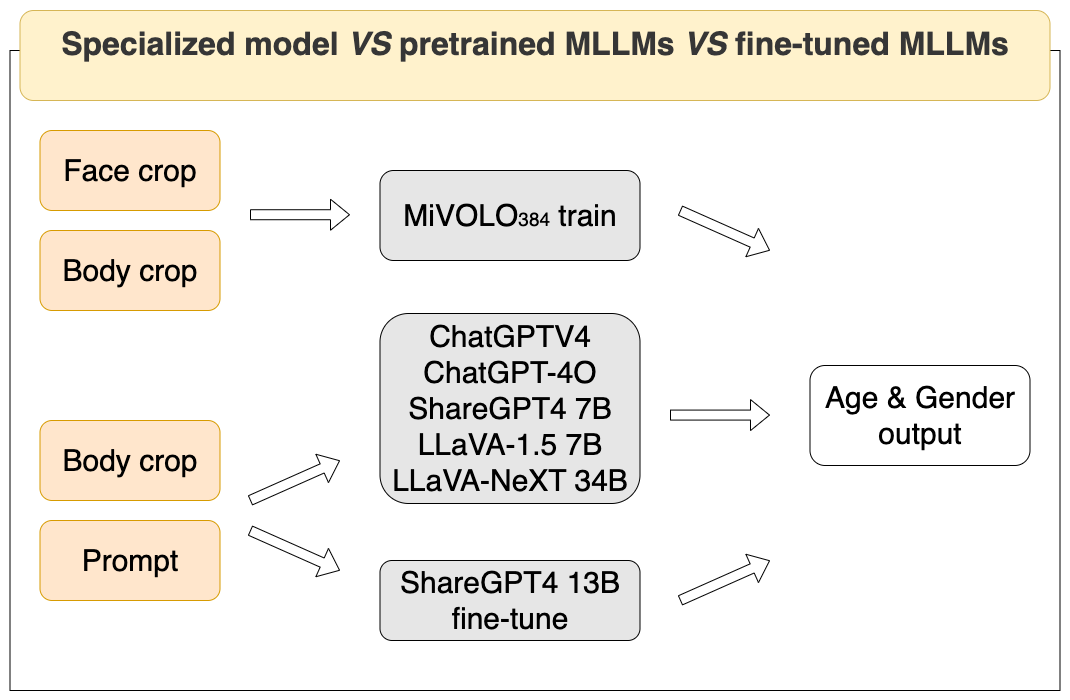}
\caption{Graphical representation of the study focused on comparison specialized model for age and gender estimation and multimodal models.}\label{fig:overview}
\end{figure}

Figure \ref{fig:example} demonstrates an example of work of evaluated models and figure \ref{fig:overview} provides a graphical representation.

\section{Related Works} \label{section:related_work}

\noindent\textbf{Age and gender estimation models.} Different researchers approach the problem of recognizing a person's age in various ways and typically address it using classical machine learning methods, CNNs or transformer-based models, primarily relying on face crops as input data.

Authors of \cite{movingwindow,qfap,haesvm} employ classical machine learning methods to tackle the regression problem. \cite{ageregression} utilizes ResNet34 to determine age through ranking of results of multiple binary classification models.

The work \cite{agegendercnn} approaches the recognition task of facial attributes, such as age and gender, as multiclass and binary classification problems, respectively, employing CNN to generate predictions. \cite{facialrepresentations} also deals with classification problems, adapting MobileNet to simultaneously predict age and gender. \cite{facialattrrec} trains CNN to classify gender and age as part of multiple facial analysis tasks using multi-task learning. \cite{agenet} utilizes two GoogLeNet models to predict age: an age classifier and an age regressor. \cite{C3AE} represents age as a convex combination of two other numbers and employs a CNN to predict the weights for these numbers.

The publication \cite{graphage} leverages CNN and GCN \cite{gcn} to extract semantic features alongside structural information from the face image for age prediction.

Neural networks based on transformers with an attention mechanism \cite{attention}, commonly employed for various natural language processing (NLP) tasks, are also widely utilized for CV tasks. Authors of \cite{attentionpatch,HierarchicalAttention} utilize not only CNNs, but also attention mechanisms, directing the model to focus on features relevant for age estimation. \cite{attentionpatch} selects the most informative age-speciﬁc patches for age estimation. \cite{HierarchicalAttention} uses transformer to aggregate the sequence of embeddings extracted by CNN and further utilyze the aggregated feature vector for age estimation. 
MiVOLO \cite{mivolo2023} employs a transformer to estimate age and gender using face and body crops as input data. In this paper, we enhance MiVOLO, resulting in a model that outperforms all the specialized models mentioned above.

\noindent\textbf{Multimodal Models.} Pre-trained vision-language models like CLIP \cite{clip} are extensively utilized in computer vision tasks. They notably improve performance across various downstream tasks by effectively matching text and images \cite{denseclip,odclip,eiclip}. 

Some works \cite{2306.13856,ordinalclip,CZLCIAE} use the CLIP pre-trained model for age estimation through visual and text embeddings matching. 
However, vision-language models still encounter difficulties in understanding instructions, capturing context, and adapting to unseen tasks. Consequently, many researchers are investigating ways to transfer the capabilities of more powerful LLMs into the visual domain, leading to the development of multimodal large language models (MLLM) \cite{flamingo,Blip2,mplugowl,llava,minigpt4,kosmos2,visionLLM}. 
Other researchers are leveraging the robust abilities of MLLM for multimodal understanding and generation to address vision tasks~\cite{lisa,instructblip,llavamed,Video-ChatGPT,mgie}.

In this paper, we aim to evaluate the capabilities of MLLMs in age and gender estimation tasks.

Examining the potential of multimodal ChatGPT (ChatGPT4V\cite{gpt4}), authors of \cite{2401.13641} assess its aptitude in predicting various facial attributes and executing face recognition tasks. With zero training the model outperformed a specialized model in age recognition, but performed less effectively in gender classification. We also compare the ChatGPT4V\cite{gpt4} model with a specialized trained model and also consider open source state-of-the-art MLLMs such as ShareGPT4V\cite{sharegpt}, LLaVa-1.5\cite{llava}, LLaVa-NeXT\cite{liu2024llavanext}. We also fine-tune a MLLM for the task of gender and age estimation to compare it with a specialized model.

\section{Enhanced MiVOLO Model} \label{section:mivolov2}

In this section, the improvements made to the MiVOLO \cite{mivolo2023} model are briefly described to achieve state-of-the-art performance. The enhanced model is used in subsequent experiments as a benchmark to compare with MLLMs, serving as an anchor specialized model. The original model from \cite{mivolo2023} is referred to as $MiVOLO_{224}$, as it was trained with 224x224 input size. We train the MiVOLO model with 384x384 input size on the extended train dataset. The enhanced model is referred to as $MiVOLO_{384}$ accordingly.

\subsection{Evaluation Metrics}

The same evaluation metrics as in the original study \cite{mivolo2023} were utilized:

\begin{itemize}
    \item Mean Absolute Error (\textbf{MAE}) for age estimation.
    \item Cumulative Score at 5 (\textbf{CS@5}) for age estimation.
    \item \textbf{Accuracy} for gender prediction.
\end{itemize}

Additionally, the Mean Average Percentage Error (\textbf{MAPE}) was slightly modified in this study as follows:
\begin{equation}
\mbox{MAPE} = \frac{1}{n}\sum_{i=1}^n \left|\frac{y^{pred}_i-y^{gt}_i}{y^{gt}_i + \epsilon}\right|
\end{equation}
An $\epsilon=1$ was chosen in the denominator for two main reasons: to prevent division by zero and to mitigate excessively high percentage errors in cases involving infants. For instance, employing $\epsilon=0.083$ (approximately one month) would result in disproportionately large errors for infants, thereby significantly biasing the MAPE.

\subsection{Datasets}

The same data as in the MiVOLO paper \cite{mivolo2023} was used, with an extension of the training dataset by approximately 40\%, resulting in over 807,694 samples: 390,730 images of males and 416,964 images of females. The extended version of LAGENDA\cite{mivolo2023} train dataset is referred to as $LAGENDA_{ext}$. The extension was achieved primarily through production pipelines and supplemented with open-source data, such as LAION-5B \cite{laion}. Focus was given to selecting images where the original $MiVOLO_{224}$ model made significant errors. Additionally, efforts were made to balance the distribution's right tail, as the original training dataset was imbalanced for ages above 70 years. The test LAGENDA dataset was taken unchanged.

\subsection{Experiment Details}

Many experiments were conducted with additional training stage image augmentations, but only one new augmentation face blurring was retained, to imitate social network filters or effects from smartphone cameras. In the first stage of training, where a single-input model that uses only faces was trained, this blur was applied slightly, with a 5\% random chance. In the second stage, with double-input, blurring parameters were significantly increased, with up to a 70\% probability. The model was also trained with a 384x384 input size instead of the originally used 224x224, showing much better results for the in-the-wild domain. Dropout and drop-path rates were decreased to 0.1, due to the large and diverse dataset.

\begin{table*}[h!]
\centering
\begin{tabular}{|l|l|c|c|c|c|}
\hline
 \multicolumn{1}{|c}{Model} & \multicolumn{1}{|c|}{Train Dataset} & \multicolumn{1}{c|}{Test Dataset} & \multicolumn{1}{c|}{MAPE, \% $\downarrow$} & \multicolumn{1}{c|}{MAE $\downarrow$} & \multicolumn{1}{c|}{CS@5, \% $\uparrow$} \\ [1ex] 
\hline
FP-Age \cite{fpage} & IMDB-clean & IMDB-clean & - & 4.68 & 63.78 \\[0.5ex] 
\hline
$MiVOLO_{224}$\cite{mivolo2023} & LAGENDA & IMDB-clean & 11.43 & 4.09 & 69.72 \\
 & & LAGENDA & 13.19 & 3.99 & 71.27 \\[0.5ex] 
\hline
$MiVOLO_{384}$ & $LAGENDA_{ext}$ & IMDB-clean & \textbf{11.01} & \textbf{3.97} & \textbf{71.16} \\
 & & LAGENDA & \textbf{12.06} & \textbf{3.65} & \textbf{74.48} \\[0.5ex] 
\hline
\end{tabular}
\caption{Comparison of $MiVOLO_{224}$ and $MiVOLO_{384}$. Age performance in face + body mode.}
\label{table:mivolov2_face_body_results}
\end{table*}

\begin{table*}[h!]
\centering
\begin{tabular}{|l|l|c|c|c|c|}
\hline
 \multicolumn{1}{|c}{Model} & \multicolumn{1}{|c|}{Train Dataset} & \multicolumn{1}{c|}{Test Dataset} & \multicolumn{1}{c|}{MAPE, \% $\downarrow$} & \multicolumn{1}{c|}{MAE $\downarrow$} & \multicolumn{1}{c|}{CS@5, \% $\uparrow$} \\ [1ex] 
\hline
$MiVOLO_{224}$\cite{mivolo2023} & LAGENDA & IMDB-clean & 19.25 & 6.66 & 47.53 \\
 & & LAGENDA & 28.88 & 7.41 & 49.64 \\[0.5ex] 
\hline
$MiVOLO_{384}$ & $LAGENDA_{ext}$ & IMDB-clean & \textbf{17.40} & \textbf{6.03} & \textbf{52.11} \\
 & & LAGENDA & \textbf{24.64} & \textbf{6.16} & \textbf{55.90} \\[0.5ex] 
\hline
\end{tabular}
\caption{Comparison of $MiVOLO_{224}$ and $MiVOLO_{384}$. Age performance in body only mode. LAGENDA in the train column refers to a part of the dataset used for training, as described in \cite{mivolo2023}.}
\label{table:mivolov2_body_only_results}
\end{table*}

\begin{table*}[h!]
\centering
\begin{tabular}{|l|l|c|c|}
\hline
 \multicolumn{1}{|c}{Model} & \multicolumn{1}{|c|}{Train Dataset} & \multicolumn{1}{c|}{Test Dataset} & \multicolumn{1}{c|}{Gender Acc, \% $\uparrow$} \\ [1ex] 
\hline
FP-Age \cite{fpage} & IMDB-clean & IMDB-clean & - \\[0.5ex] 
\hline
$MiVOLO_{224}$\cite{mivolo2023} & LAGENDA & IMDB-clean & 99.55 \\
 & & LAGENDA & 97.36 \\[0.5ex] 
\hline
$MiVOLO_{384}$ & $LAGENDA_{ext}$  & IMDB-clean & \textbf{99.68} \\
 & & LAGENDA & \textbf{97.99} \\[0.5ex] 
\hline
\end{tabular}
\caption{Comparison of $MiVOLO_{224}$ and $MiVOLO_{384}$. Gender Accuracy.}
\label{table:mivolov2_gender_acc_results}
\end{table*}

\begin{table}[h!]
\centering
\begin{tabular}{|l|c|c|}
\hline
 \multicolumn{1}{|c|}{Model} & \multicolumn{1}{c|}{Age MAE $\downarrow$} & \multicolumn{1}{c|}{Age CS@5, \% $\uparrow$} \\ [1ex] 
\hline
ResNet-50 \cite{2307.04570} & 3.96 & - \\[0.5ex] 
\hline
$MiVOLO_{224}$\cite{mivolo2023} \cite{mivolo2023} & 4.09 & 70.73 \\[0.5ex] 
\hline
$MiVOLO_{384}$ & \textbf{3.89} & \textbf{73.26} \\[0.5ex] 
\hline
\end{tabular}
\caption{Comparison of models using the CACD test split. MiVOLO models are evaluated in face + body mode.}
\label{table:cacd_results}
\end{table}

\begin{table}[h!]
\centering
\begin{tabular}{|l|c|c|}
\hline
 \multicolumn{1}{|c|}{Model} & \multicolumn{1}{c|}{Age Acc, \% $\uparrow$} & \multicolumn{1}{c|}{Gender Acc, \% $\uparrow$} \\ [1ex] 
\hline
$MiVOLO_{224}$ \cite{mivolo2023} & 61.07 & 95.73 \\[0.5ex] 
\hline
$MiVOLO_{384}$ & \textbf{62.28} & \textbf{97.5} \\[0.5ex] 
\hline
\end{tabular}
\caption{Comparison of models using the FairFace validation margin125 split. MiVOLO models are evaluated in face + body mode.}
\label{table:fairface_results}
\end{table}

\subsection{Results}

Tables \ref{table:mivolov2_face_body_results}, \ref{table:mivolov2_body_only_results} show comparison of original $MiVOLO_{224}$ and $MiVOLO_{384}$ models. The results for $MiVOLO_{384}$ establish new state-of-the-art results for specialized models. For comparisons with MLLM models, see the following section. Tables \ref{table:cacd_results}, \ref{table:fairface_results} provide a comparison of results using the Cross-Age Celebrity Dataset (CACD2000) \cite{cacd} test split and FairFace \cite{fairface} validation `margin125` split.

\section{MLLM Models vs. Specialized Model}
\label{section:mllm}

In this section, we compare the capabilities of MLLMs with MiVOLO in age and gender estimation tasks on various benchmarks. Additionally, we investigate the effects of fine-tuning MLLMs on large target datasets to enhance their accuracy in these specific tasks.

\subsection{Benchmarks} \label{section:mllm_data}
A multitude of benchmarks is available for age or gender estimation. In this study, we concentrated on those offering full-body images, when possible, and containing both age and gender labels.

Hence, we selected two datasets:
\begin{itemize}
\item IMDB-clean dataset \cite{imdbwikiclean}, which was used in the MiVOLO article \cite{mivolo2023}. The original IMDB-clean dataset was enhanced with body bounding boxes associated with facial pairs. It comprises 183,886 training images, 45,971 validation images, and 56,086 test images. We use only the test images for our evaluation.
\item LAGENDA benchmark \cite{mivolo2023}, a dataset well-balanced in terms of age and gender attributes, features face-body pairs with ground truth obtained through human annotations via weighted voting. It includes proprietary training and validation parts, and an open-source test part, containing 67,159 images from the Open Images Dataset \cite{oid4} featuring 84,192 individuals aged from 0 to 95.
\end{itemize}

These datasets are of exceptionally high quality and exhibit significant diversity.

Initially, the OpenAI API imposed a limit of 100 requests per day for the gpt-4-vision-preview models, which has since been increased to 1,500 requests per day. Due to this limitation, we randomly selected a small subset from LAGENDA \cite{mivolo2023} to run with ChatGPT4V and included it in our comparison.
We selected 200 random samples for each age group at intervals of 5 years (e.g., 0-5, 5-10, etc.) for ages $[0; 90]$. 
However, $> 21\%$ of these images had to be removed because ChatGPT refused to provide answers. As a result, this dataset contains 3,062 samples. We refer to this dataset as NanoLAGENDA.

We opted for LAGENDA over IMDB to minimize the risk that MLLMs would provide correct answers not through age and gender estimation but because of its familiarity with famous individuals, well-known movies, etc. 
On the other hand, LAGENDA, annotated by human annotators, does not have actual ground truths for labels. Nevertheless, we chose it because the risk of this drawback is lower.

Additionally, to slightly offset the drawback of annotated ages, we compiled a very small dataset of 104 samples from social networks like Instagram, primarily with real ground truth answers (we know the actual ages) for ages $[0; 105]$ and very challenging samples, which typically result in large human prediction errors. This dataset is entirely manual and is intended solely to double-check our conclusions. We refer to this small set as the Wild104 dataset. We cannot publish it because we do not own the photos.

We also used the Adience \cite{adience} benchmark to compare specialized MiVOLO models with the multimodal approaches mentioned in section \ref{section:related_work} and our fine-tuned models \ref{section:llava_finetune}. The dataset consists of 26,580 facial images. Annotations include age labels from eight age group classes and labels for the binary gender classification task.

\subsection{Method} \label{section:mllm_method}

\begin{table*}
\centering
\begin{tabular}{|l|l|c|c|c|c|c|}
\hline
\multicolumn{1}{|c|}{Base Model} & \multicolumn{1}{c|}{Body or Face Crop} & \multicolumn{2}{c|}{Age} & \multicolumn{1}{c|}{Gender} \\ \cline{3-5}
 & & MAE $\downarrow$ & CS@5, \% $\uparrow$ & Acc, \% $\uparrow$ \\ [1ex]
\hline
LLaVa-v1.5-7b & Body & 4.52 & 69.84 & 99.06 \\[0.3ex] 
\hline
ShareGPT4V-7b 0.4 epoch & Body & \textbf{3.87} & \textbf{75.93} & 99.45 \\[0.3ex]
\hline
ShareGPT4V-7b 1 epoch & Body & 3.94 & 75.34 & 99.53 \\[0.3ex]
\hline
ShareGPT4V-13b & Body & 3.93 & 75.42 & \textbf{99.54} \\[0.3ex] 
\hline
ShareGPT4V-7b & Face & 4.32 & 72.98 & 97.56 \\[0.3ex] 
\hline
\end{tabular}
\caption{Comparison of fine-tuned MLLMs on LAGENDA benchmark.}
\label{table:MLLM_finetuned_models}
\end{table*}

We used the same prompt for all models. However, the full version was necessary only for ChatGPT; for the sake of an honest comparison, we applied the same for open-source models as well. In our tests, this did not influence the answers.

\begin{tcolorbox}[title=\textbf{Prompt for MLLMs}]
Please provide the perceived gender and estimated age of the individual in this photo in the format '[gender; age]', where the age must be a single integer between 0 and 130. It should not be a range. The gender must be one of two options: 'male' or 'female'. I understand that your response will be based on visual analysis and may not accurately reflect the actual age or gender identity of the individual.
\end{tcolorbox}

A typical answer looks like:
[female; 40]

We set the temperature for all models to 0.0. For ChatGPT, we additionally set the parameter \textbf{seed} to 1234 and \textbf{n} to 1. The latter is necessary due to reports that just the seed and temperature are not sufficient to ensure deterministic and reproducible results for vision model via API.

However, for different models, a zero temperature setting can lead to various issues due to the nature of LLMs.
ChatGPT4V might occasionally fail to provide an answer, typically for queries that are either difficult or do not pass the safety system checks. We decided to remove such samples, although it gives ChatGPT4V a slight advantage.
LLaVA \cite{llava}, LLaVA-NeXT \cite{llava_improved}, and LLaVA-based ShareGPT4V \cite{sharegpt} may sometimes return an age range instead of a specific age. Since we have sampling disabled and the temperature is set to 0.0, we cannot workaround this; thus, we take the midpoint of the range in such cases. However, the number of such samples is very low, and this does not significantly impact the results.

For all models, we attempted to use the maximum possible resolution with the goal to measure the maximum possible performance without taking into accound speed of inference. Thus:
\begin{itemize}
    \item For ChatGPT, we used full-sized original crops to allow the model to split them into tiles (if large enough), which gives ChatGPT another slight advantage.
    \item For LLaVA 1.5 and ShareGPT4V, we used the original 336x336 resolution.
    \item For LLaVA-NeXT, we utilized the new Dynamic High Resolution technique, although its maximum resolution is still restricted to 672x448 or 448x672.
\end{itemize}  

\begin{table*}
\centering
\begin{tabular}{|l|l|c|c|c|}
\hline
 \multicolumn{1}{|c|}{Model} & \multicolumn{1}{|c|}{Input} & \multicolumn{1}{c|}{Age Acc, \% $\uparrow$} & \multicolumn{1}{c|}{Age MAE, \% $\downarrow$} & \multicolumn{1}{c|}{Gender Acc, \% $\uparrow$} \\ [1ex] 
\hline
OridinalCLIP \cite{ordinalclip} & Face & 61.2 & 0.47 & - \\[1ex] 
\hline
L2RCLIP \cite{2306.13856} & Face & 66.2  & 0.36 & - \\[1ex] 
\hline 
$MiVOLO_{224}$ \cite{mivolo2023} & Body \& face & 68.68 & 0.345 & 96.5 \\[0.5ex] 
\hline
$MiVOLO_{384}$ & Body \& face & \textbf{69.43} & \textbf{0.333} & \textbf{97.39} \\[0.5ex] 
\hline
ShareGPT4V 7B FT & Body & 66.7 & 0.349 & 95.65 \\[0.5ex] 
\hline
ShareGPT4V 7B FT & Face & 67.95 & 0.338 & 96.63 \\[0.5ex] 
\hline
\end{tabular}
\caption{Comparison of models using Adience benchmark. MAE here is calculated on classification labels. FT denotes models that are fine-tuned version on the corresponding input.}
\label{table:adience_results}
\end{table*}
\begin{table*}[h]
\centering
\begin{tabular}{|l|l|c|c|c|c|c|}
\hline
\multicolumn{1}{|c|}{Model} & \multicolumn{1}{c|}{Input} & \multicolumn{3}{c|}{Age} & \multicolumn{1}{c|}{Gender} \\ \cline{3-6}
 &  & MAPE, \% $\downarrow$ & MAE $\downarrow$ & CS@5, \% $\uparrow$ & Acc, \% $\uparrow$ \\ [1ex]
\hline
LLaVA 1.5 7B \cite{llava} & Entire body & 16.86 & 7.59 & 48.79 & 99.38 \\[0.5ex] 
 & W/o face & 43.18 & 18.44  & 25.71 & 94.85 \\[0.5ex] 
\hline
LLaVA-NeXT 34B\cite{liu2024llavanext} & Entire body & 13.73 & 6.20 & 55.40 & 99.51 \\[0.5ex] 
 & W/o face & 23.03 & 9.58  & 39.67 & 97.82 \\[0.5ex] 
\hline
ShareGPT4V 7B \cite{sharegpt} & Entire body & 16.71 & 7.16 & 53.24 & 99.44 \\[0.5ex] 
 & W/o face & 25.80 & 11.16  & 39.07 & 97.24 \\[0.5ex] 
\hline
ChatGPT4V \cite{gpt4} & Entire body & 12.12 & 4.66  & 68.10 & 98.43 \\[0.5ex] 
 & W/o face & 22.56 & 7.82  & 49.15 & 93.02 \\[0.5ex] 
\hline
ChatGPT4O \cite{gpt4} & Entire body & \textbf{10.42} & \textbf{4.07}  & \textbf{73.91} & 98.66 \\[0.5ex] 
 & W/o face & \underline{\textit{15.39}} & \underline{\textit{5.73}} & \underline{\textit{60.17}} & 96.92 \\[0.5ex] 
\hline
$MiVOLO_{384}$ & Body \& face & 11.61 & 4.33  & 69.90 & 97.71 \\[0.5ex] 
 & W/o face & 21.82 & \underline{\textit{7.19}} & 49.14 & 95.61 \\[0.5ex] 
\hline
ShareGPT4V 7B FT & Entire body & 10.95 & 4.22 & 72.09 & \textbf{99.51} \\[0.5ex] 
 & W/o face & 20.53 & 7.64 & 49.95 & 97.48 \\[0.5ex] 
\hline
ShareGPT4V 13B FT & Entire body & 11.30 & 4.26 & 73.34 & 99.44 \\[0.5ex] 
 & W/o face & 19.90 & 7.40 & 54.01 & \underline{\textit{98.10}} \\[0.5ex] 
\hline
\end{tabular}
\caption{Comparison of performance on NanoLAGENDA benchmark. Models are evaluated with different type of input information. \textbf{Bold} indicates the best model performance running with all available information about the person. \underline{\textit{Underline}} shows the best performance running without faces.}
\label{table:nanolagenda}
\end{table*}

\subsection{LLaVA Finetune} \label{section:llava_finetune}

In this section, we explore the fine-tuning of a general-purpose multimodal network, namely LLaVA, for age and gender recognition tasks. 

Building on insights from the MiVOLO \cite{mivolo2023}, simultaneous training for both tasks has shown to be advantageous. Consequently, our training approach mirrors the evaluation methodology described in the preceding section.

Initially, the only available version for research was LLaVa-v1.5's training code in open source, guiding our choice of starting point. Data conversion utilized the same prompt as for ChatGPT, employing a single crop as the input image. Various experiments using both whole body and face crops were conducted, with whole body crops yielding marginally superior results for gender and age recognition tasks.

Further, ShareGPT4V, which is derived from LLaVa code and demonstrates slightly improved metrics, became the source of checkpoints from the Hugging Face hub, used with minor modifications to the LLaVa code. The ShareGPT4V training code was not accessible at the time.

Another experimental direction focused on training for direct gender classification and age prediction, rather than text prediction, using linear layers along with MSE and cross-entropy losses, paralleling the approach used for LLaMa sequence classification tasks \cite{llama}. This approach is not mentioned in the results due to bad performance.

The optimal hyperparameters identified through our experiments are as follows:

\begin{center}
\begin{tabular}{cc}
    \textbf{Hyperparameter} & \textbf{Value} \\
    \hline
    Learning rate & $2 \times 10^{-6}$ \\
    LoRA & disabled (full fine-tuning) \\
    Per-device train batch size & 32 \\
    Number of train epochs & 1 \\
    Checkpoint every & 450 iterations \\
    Warmup ratio & 0.03 \\
    LR scheduler type & cosine \\
\end{tabular}
\end{center}

Table \ref{table:MLLM_finetuned_models} provides a comparative overview of the fine-tuned MLLMs. Note that the best-performing model in Table \ref{table:MLLM_finetuned_models} (ShareGPT4V-7b 0.4 epoch) is referred to as ShareGPT4V 7B fine-tuned in subsequent sections, representing the optimal checkpoint achieved.

The finest results were obtained using the ShareGPT4V-7b model trained with whole body crops. Notably, the best metrics were observed at the 900th iteration checkpoint, approximately 40\% through one epoch, suggesting an early stop with low learning rate strategy might be beneficial and shows that MLLMs, as generalist models, are easy to fine-tune with smaller data amount. 

Important to mention, that after a few iterations, the training loss stabilizes at around 0.32, and further training steps may lead to overfitting. This is corroborated by the observation that later iterations yield slightly diminished results. Additionally, extended training may impair the model's assistant capabilities, restricting responses to the trained format. This limitation could potentially be mitigated by diversifying the training data beyond age estimation tasks. It is also worth noting that testing on the LAGENDA test set requires approximately 2.5 hours on 8 A6000 NVidia GPUs, a significant duration relative to the training time of about 11 hours for one epoch. While other iteration intervals may yield superior results, our study focused on evaluating every 450th iteration to optimize training time and costs.

Future developments could explore dual-crop inputs (body + face as separate images), as seen in the original $MiVOLO_{224}$ model. However, the feasibility of training existing models with multiple images per input remains an open question.

\subsection{Results} \label{section:comparison_results}

The table \ref{table:adience_results} presents a comparison of specialized MiVOLO models and multimodal approaches using the Adience benchmark. Following the methodology of \cite{mivolo2023}, we mapped regression predictions to the nearest intervals (classes).

Table \ref{table:nanolagenda} displays analogous results for a randomly sampled subset of NanoLAGENDA, including evaluations of ChatGPT4V.

\begin{figure*}[!h]
\centering
\includegraphics[width=12cm]{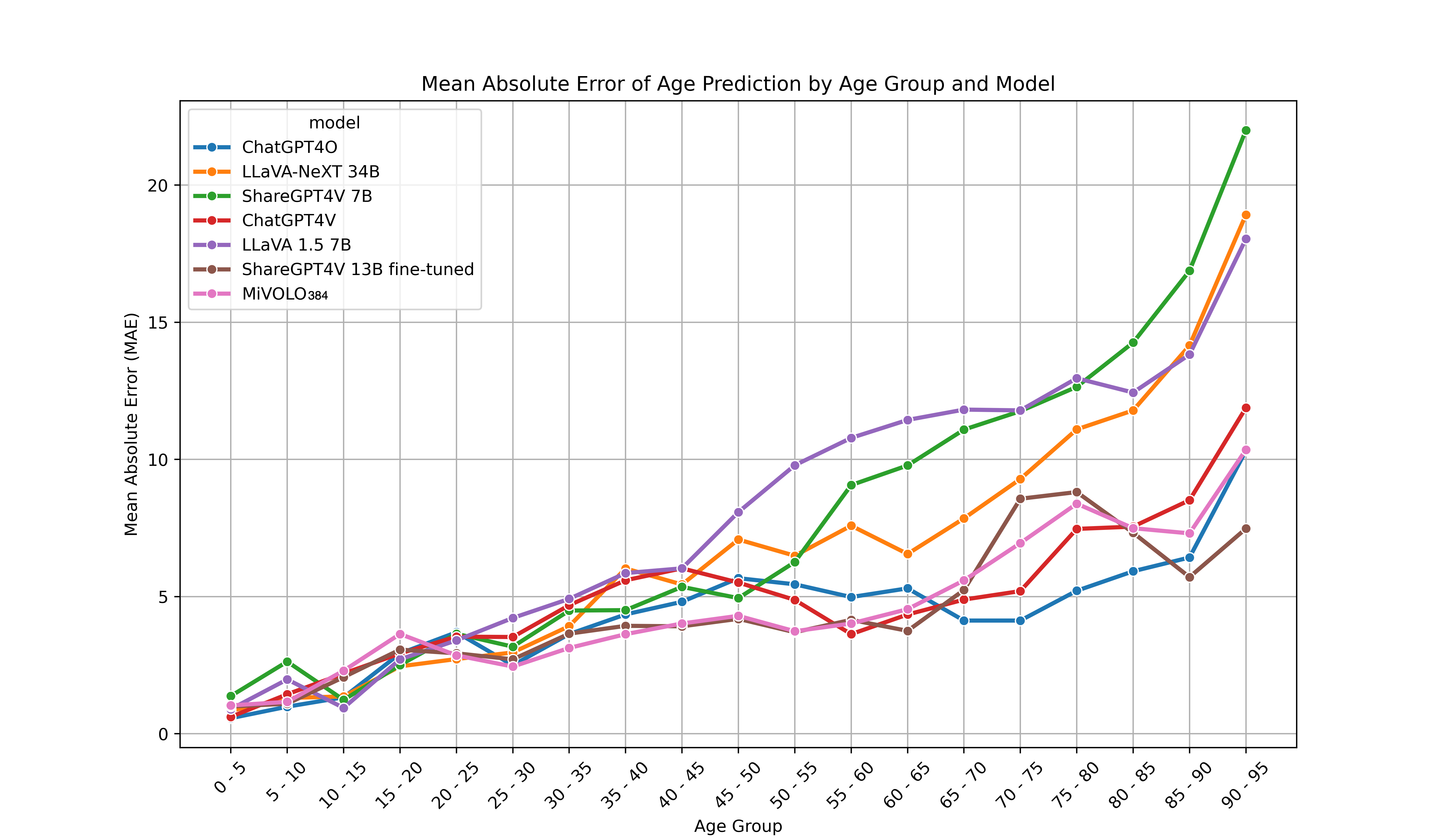}\\
\caption{Relationship between MAE and age group across different models tested on the NanoLAGENDA benchmark.}
\label{figure:mae_figure}
\end{figure*}

Figure \ref{figure:mae_figure} illustrates the relationship between MAE and age for NanoLAGENDA, with age intervals set at 5-year steps. Interestingly, ChatGPT's performance closely parallels that of models trained specifically on our dataset, particularly underperforming in the 25 to 55 age range — a notably common age group in the dataset.

Table \ref{table:imdb_and_lagenda} reports outcomes for the full-sized LAGENDA and IMDB datasets for applicable models.

Table \ref{table:nano_in_the_wild} shows results for the Wild104 benchmark, reaffirming prior findings with real-world ground truth labels and highlighting a different visual domain.

\begin{table*}[h]
\centering
\begin{tabular}{|l|c|c|c|c|c|}
\hline
\multicolumn{1}{|c|}{Model} & \multicolumn{1}{c|}{Test Dataset} & \multicolumn{3}{c|}{Age} & \multicolumn{1}{c|}{Gender} \\ \cline{3-6}
 &  & MAPE, \% $\downarrow$ & MAE $\downarrow$ & CS@5, \% $\uparrow$ & Acc, \% $\uparrow$ \\ [1ex]
\hline
$MiVOLO_{384}$ & IMDB-clean & \textbf{11.01} & \textbf{3.97} & \textbf{71.16} & \textbf{99.68} \\
 &  LAGENDA & 12.06 & 3.65 & 74.48 & 97.99 \\[0.5ex] 
\hline
LLaVA-NeXT 34B & IMDB-clean & 16.04 & 5.66 & 59.77 & 99.15 \\
vanilla \cite{liu2024llavanext} & LAGENDA & 16.97 & 5.19 & 62.17 & \textbf{99.47} \\[0.5ex] 
\hline
ShareGPT4V 7B & IMDB-clean & 12.07 & 4.40 & 70.28 & 99.47 \\
fine-tuned &  LAGENDA & \textbf{11.47} & \textbf{3.52} & \textbf{79.66} & 99.44 \\[0.5ex] 
\hline
\end{tabular}
\caption{Comparison of performance on IMDB and LAGENDA benchmarks.}
\label{table:imdb_and_lagenda}
\end{table*}

\begin{table*}[!h]
\centering
\begin{tabular}{|l|c|c|c|c|c|}
\hline
\multicolumn{1}{|c|}{Model} & \multicolumn{3}{c|}{Age} & \multicolumn{1}{c|}{Gender} \\ \cline{2-5}
 & MAPE, \% $\downarrow$ & MAE $\downarrow$ & CS@5, \% $\uparrow$ & Acc, \% $\uparrow$ \\ [1ex]
\hline
ChatGPT4V \cite{gpt4} & 19.95 & 7.07 & 48.08 & 91.35 \\[0.5ex] 
\hline
ChatGPT4O \cite{gpt4} & \textbf{16.11} & \textbf{6.07} & \textbf{58.65} & 91.35 \\[0.5ex] 
\hline
LLaVA-NeXT 34B vanilla \cite{liu2024llavanext} & 22.16 & 9.23 & 42.31 & 96.15 \\[0.5ex] 
\hline
$MiVOLO_{384}$ & 19.82 & 6.26 & 53.67 & 96.17 \\[0.5ex] 
\hline
ShareGPT4V 7B fine-tuned & 19.36 & 7.01 & 53.85 & 95.19 \\[0.5ex] 
\hline
ShareGPT4V 13B fine-tuned & 18.27 & 6.79 & 57.69 & \textbf{97.12} \\[0.5ex] 
\hline
\end{tabular}
\caption{Comparison of performance on the Wild104 benchmark.}
\label{table:nano_in_the_wild}
\end{table*}

\begin{figure*}
\centering
\includegraphics[width=13cm, height=6.5cm]{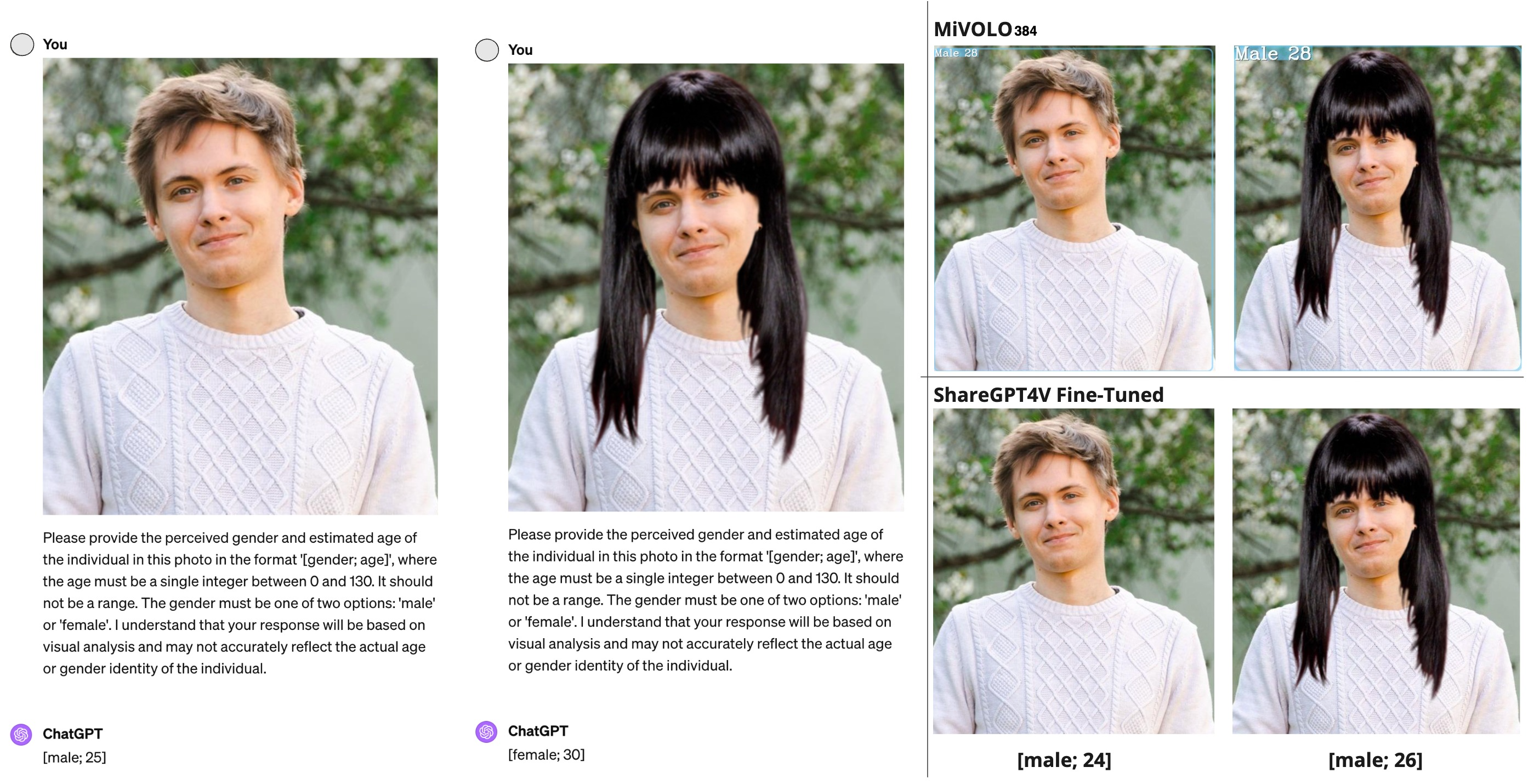}
\caption{Failure case: Through manual analysis of ChatGPT's gender misclassification, we have discovered that it sometimes makes mistakes with examples where men have long hair. This is a humorously artificial illustrative example of such failure cases. Both MiVOLO and the fine-tuned as well as vanilla LLaVA models remain stable.}\label{fig:fail_case}
\end{figure*}
\section{Conclusions}

This study aimed to assess the efficacy of cutting-edge specialized models in comparison to MLLMs for age and gender estimation tasks.

Our findings reveal a nuanced view. MLLMs, despite not being explicitly trained for facial or bodily analyses to deduce personal attributes, exhibit exceptional capabilities. Notably, models such as ChatGPT stand out by harnessing vast amounts of visual information and training on extensive datasets, demonstrating significant proficiency in tasks beyond their original design. 
Our analysis identified ChatGPT-4O as the most precise MLLM for age estimation across numerous benchmarks, despite encountering challenges such as occasional refusal to process images with significant losses ($>21\%$ for our data) and the need for Dynamic High Resolution, which demands a much higher budget. While this comparison may not be entirely equitable, it highlights the model's capabilities in 'maximum power mode'. Among open-source alternatives, LLaVA-NeXT 34B leads in this area. 
At the same time, the improved specialized model $MiVOLO_{384}$ surpasses all general-purpose open-source MLLMs in age estimation. However, for certain data segments and metrics, fine-tuned specialized versions of LLaVA prove more effective. Such fine-tuned MLLMs present a promising solution for many tasks where computational cost is not a primary concern. Compared to the tricky and expert-driven training required for MiVOLO, fine-tuning an MLLM is considerably simpler, requiring only the same dataset as the specialized model and minimal expertise. Original hyperparameters and losses can be used.

The study highlights the superior performance of MLLMs also in gender identification tasks even without any fine-tuning, surpassing that of specialized models. This emphasizes the significance of high-level feature recognition and contextual understanding in this task, where nearly all MLLMs excel. However, ChatGPT-4V and even the newest ChatGPT-4O stand out for their subpar performance, possibly due to an overemphasis on certain features like hair, which might be influenced by its training data or safety mechanisms. For visualizations, refer to Figure \ref{fig:fail_case}. The opaque nature of its development process hampers definitive conclusions.

Overall, our research indicates that with minor adjustments, open-source MLLMs can achieve or even surpass the performance of specialized models, suggesting a potential shift towards versatile, general-purpose networks in computer vision. The flexibility of language models offers significant advantages for a wide range of applications, especially in scenarios where computational resources and inference speed are not primary concerns. However, it is important to note that, for the time being, the computational cost of MLLMs cannot be directly compared to that of specialized models — the difference can span thousands of times. Possibly, in the future, Multimodal Tiny Language Models could turn the tables.

\bibliographystyle{splncs04}
\bibliography{main}

\section*{Authors}
\noindent {\bf Maksim Kuprashevich} received a Bachelor's degree in Computer Science from the Saint-Petersburg State Institute of Technology. He is currently a Research Team Lead. His research interests include deep learning, particularly in vision, language, and generative models.\\

\noindent {\bf Grigorii Alekseenko} received a Specialist degree in Fundamental Mathematics and Mechanics from Moscow State University. He is currently a Data Scientist. His research interests include computer vision, multimodality, and diffusion neural networks. \\

\noindent {\bf Irina Tolstykh} received a Bachelor's degree in Fundamental Informatics and Information Technology from Saint-Petersburg State University. She is currently a Senior Data Scientist. Her research interests in machine learning include applying deep learning methods to computer vision and natural language processing, as well as exploring generative AI. \\

\end{document}